\documentclass[a4paper,twoside]{article}

\usepackage{subcaption}
\usepackage{calc}
\usepackage{amssymb}
\usepackage{amstext}
\usepackage{amsmath}
\usepackage{amsthm}
\usepackage{multicol}
\usepackage{pslatex}
\usepackage{apalike}
\usepackage{algorithm2e}
\usepackage[bottom]{footmisc}
\usepackage{style}    
\usepackage{graphicx}  
\pdfoutput=1 

\begin{document}

\title{SPNeRF: Open Vocabulary 3D Neural Scene Segmentation with Superpoints}

\author{\authorname{Weiwen Hu\sup{1}, Niccolò Parodi\sup{1,2}, Marcus Zepp\sup{1}, Ingo Feldmann\sup{1}, Oliver Schreer\sup{1}, Peter Eisert\sup{1,3}}
\affiliation{\sup{1} Fraunhofer Heinrich Hertz Institute, Berlin, Germany}
\affiliation{\sup{2} Technische Universität Berlin, Germany}
\affiliation{\sup{3} Humboldt-Universität zu Berlin, Germany}
}


\abstract{Open-vocabulary segmentation, powered by large visual-language models like CLIP, has expanded 2D segmentation capabilities beyond fixed classes predefined by the dataset, enabling zero-shot understanding across diverse scenes. Extending these capabilities to 3D segmentation introduces challenges, as CLIP’s image-based embeddings often lack the geometric detail necessary for 3D scene segmentation. Recent methods tend to address this by introducing additional segmentation models or replacing CLIP with variations trained on segmentation data, which lead to redundancy or loss on CLIP's general language capabilities. To overcome this limitation, we introduce SPNeRF, a NeRF based zero-shot 3D segmentation approach that leverages geometric priors. We integrate geometric primitives derived from the 3D scene into NeRF training to produce primitive-wise CLIP features, avoiding the ambiguity of point-wise features. Additionally, we propose a primitive-based merging mechanism enhanced with affinity scores. Without relying on additional segmentation models, our method further explores CLIP’s capability for 3D segmentation and achieves notable improvements over original LERF.}

\onecolumn \maketitle \normalsize \setcounter{footnote}{0} \vfill

\section{\uppercase{Introduction}}
\label{sec:introduction}

Traditional segmentation models are often limited by their reliance on closed-set class definitions, which restricts their applicability to dynamic real-world environments, where new and diverse objects frequently appear. Open-vocabulary segmentation, powered by large visual-language models (VLMs), such as CLIP \cite{clip}, overcomes this barrier by enabling zero-shot recognition of arbitrary classes based on natural language queries. This adaptability is crucial in applications like autonomous navigation, augmented reality, and robotic perception, where it is impractical to exhaustively label every possible object. CLIP aligns 2D visual and language features within a shared embedding space, enabling image classification/understanding without the need for extensive retraining.

In 2D segmentation, this flexibility has led to the development of powerful models \cite{segclip,groupvit}. Some methods, like OpenSeg \cite{openseg} and LSeg \cite{lseg}, leverage CLIP’s embeddings and additional segmentation annotation to perform dense, pixel-wise 2D segmentation. These methods have demonstrated that open-vocabulary segmentation not only outperforms traditional closed-set models in adaptability but also provides a scalable solution for handling diverse tasks across various domains. However, transitioning from 2D to 3D segmentation introduces unique challenges, as 3D environments require neural models to interpret complex spatial relationships and geometric structures that 2D models do not address.

To tackle these challenges, recent works such as LERF \cite{lerf2023} have embedded CLIP features within 3D representations like Neural Radiance Fields (NeRF) \cite{mildenhall2020nerf}. These methods aim to bridge 2D VLMs with 3D scene understanding by enabling open-vocabulary querying across 3D spaces. However, due to the image-based nature of CLIP embeddings, which often lack the geometric precision required for fine-grained 3D segmentation, methods either struggle with segmentation in complex scenes \cite{lerf2023}, or integrate additional segmentation models \cite{engelmann2024opennerf,takmaz2023openmask3d}.

To address these limitations, we propose SPNeRF, a NeRF-based approach specifically designed to incorporate geometric priors directly from the 3D scene. Unlike prior methods that rely solely on CLIP’s image-centric features, SPNeRF leverages geometric primitives to enhance segmentation accuracy. By partitioning the 3D scene into geometric primitives, SPNeRF creates primitive-wise CLIP embeddings that preserve geometric coherence. This enables the model to better align CLIP’s semantic representations with the underlying spatial structure, mitigating the ambiguities often associated with point-wise features.

Furthermore, SPNeRF introduces a merging mechanism for these geometric primitives, incorporating an affinity scoring system to refine segmentation boundaries. This approach allows SPNeRF to capture semantic relationships between superpoints, resulting in a more accurate and consistent segmentation output. While avoiding additional segmentation models or segmentation-specific training data, our SPNeRF provides a zero-shot architecture for 3D segmentation tasks.

The main contributions of SPNeRF are as follows:
\begin{itemize}
    \item \textit{Geometric primitives for improved 3D segmentation}: We integrate geometric primitives into NeRF for open-set segmentation, introducing a loss function that maintains consistency within primitive-wise CLIP features, ensuring coherent segmentation across 3D scenes;
    \item \textit{Primitive-based merging with affinity scoring}: SPNeRF employs a merging mechanism that uses affinity scoring to refine segmentation, capturing semantic relationships among primitives and improving boundary precision;
    \item \textit{Enhanced segmentation without additional models}: By leveraging primitive-based segmentation and affinity refinement, SPNeRF improves segmentation accuracy over LERF without relying on extra segmentation models, preserving open-vocabulary capabilities with a streamlined architecture;
\end{itemize}

\section{\uppercase{Related Work}}

\subsection{2D Vision-Language Models}

CLIP \cite{clip} has fueled the explosive growth of large vision-language models. It consists of an image encoder and a text encoder, each mapping their respective inputs into a shared embedding space. Through contrastive training on large-scale image-caption pairs, the encoders align encoded image and caption features to the same location in the embedding space if the caption accurately describes the image, otherwise the encoders push them away. 2D segmentation methods building on CLIP have extended its potential. Approaches by \cite{openseg,lseg} achieve open vocabulary segmentation by training or fine tuning on datasets with segmentation info. These datasets tend to have limited vocabulary due to expensive annotation cost of segmentation, which leads to reduced open vocabulary capacity as stated in \cite{sun2024cliprnnsegmentcountless,lerf2023}. The works of \cite{sun2024cliprnnsegmentcountless,lan2024clearclipdecomposingcliprepresentations} explore alternative approaches to maximize CLIP's potential, achieving competitive semantic segmentation results while preserving its general language capabilities.

\subsection{Neural Radiance Fields}

Neural Radiance Fields (NeRFs) \cite{mildenhall2020nerf} represent 3D geometry and appearance with a continuous implicit radiance field, parameterized by a multilayer perceptron (MLP). They also provide a flexible framework for integrating 2D-based information directly into 3D, supporting complex semantic and spatial tasks. Works, such as \cite{cen2024segment3dradiancefields}, bring class-agnostic segmentation ability from 2D foundation models to 3D. The method proposed by \cite{siddiqui2022panopticlifting3dscene} adds multiple branches to NeRF for instance segmentation. Works, like \cite{engelmann2024opennerf}, extend NeRF’s capabilities to 3D scene understanding by leveraging pixel-aligned CLIP features from 2D models like \cite{lseg}. Our work builds on LERF \cite{lerf2023}, which utilizes pyramid-based CLIP supervision for open-vocabulary 3D segmentation. However, while LERF's global CLIP features enable effective language-driven queries, they often lack the precision needed for 3D segmentation - a limitation our method seeks to improve.

\begin{figure*}[t]
  \centering
  \includegraphics[width=\textwidth]{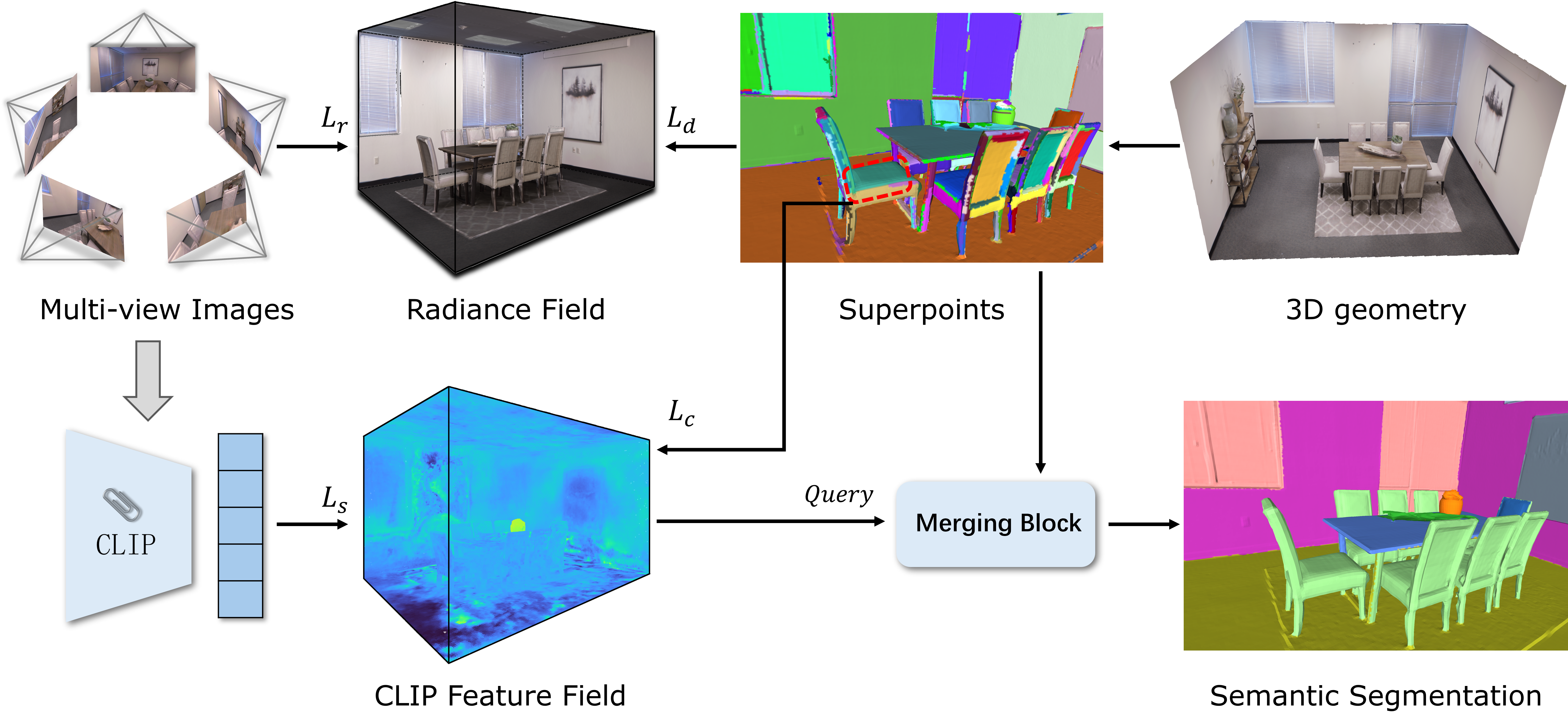}
  \caption{\textbf{Overview of SPNeRF Pipeline.} Given 2D posed images as input, SPNeRF optimizes a 3D CLIP feature field by distilling vision-language embeddings from the CLIP image encoder. Simultaneously, the radiance field is trained in parallel. Superpoints, which are extracted from the 3D geometry, are used to enhance both the radiance field and CLIP feature field during optimization, ensuring better alignment of semantic and spatial information. The training process leverages a combination of loss functions $\mathcal{L}$ to refine the consistency and accuracy of the feature representations. The merging block combines query labels with superpoints information to produce semantic segmentation results.}
  \label{fig:overview}
 \end{figure*}

\subsection{3D Open-Vocabulary Segmentation}
Extending open-vocabulary segmentation from 2D to 3D brings challenges, as 2D vision-language models like CLIP struggle with the spatial complexity of 3D scenes. Methods like OpenMask3D \cite{takmaz2023openmask3d} accumulate and average CLIP features obtained from instance-centered image crops. The features are then used to represent the 3D instance for instance segmenation.  OpenScene \cite{peng2023openscene3dsceneunderstanding} projects 2D CLIP features into 3D by aligning point clouds with 2D embeddings using a 3D convolutional network. This enables language-driven queries without labeled 3D data. Other methods, like \cite{yang2024regionplcregionalpointlanguagecontrastive}, leverage image captioning models \cite{wang2022ofaunifyingarchitecturestasks} to generate textual descriptions of images, and align point cloud features with open-text representations. Based on LERF, our method leverages NeRF as a flexible framework for 2D to 3D lifting, avoiding the geometric consistency limitation of direct 2D projection methods. Furthermore, we take advantage of simple geometric primitives instead of full 3D object masks in \cite{takmaz2023openmask3d} to enhance spatial coherence across the scene.

\section{\uppercase{Method}}
In this section, we introduce SPNeRF, our proposed method for zero-shot 3D semantic segmentation. SPNeRF extends NeRF by incorporating CLIP features into an additional feature field, building on principles similar to LERF. We outline the loss functions which are designed to train this feature field, ensuring improved consistency of CLIP features within superpoints. Furthermore, we detail a merging mechanism for robust semantic class score and leverage superpoint affinity scores to refine the segmentation results. A comprehensive overview of the SPNeRF pipeline is presented in Figure~\ref{fig:overview}.

\subsection{Preliminary: LERF}

We first introduce Language Embedded Radiance Fields (LERF) \cite{lerf2023} which SPNeRF is built upon. LERF integrates CLIP embeddings into a 3D NeRF framework, enabling open-vocabulary scene understanding by grounding semantic language features spatially across the 3D field. Unlike standard NeRF outputs \cite{mildenhall2020nerf,barron2021mipnerf}, LERF introduces a dedicated language field, which leverages multi-scale CLIP embeddings to capture semantic information across varying levels of detail. This language field is represented by $\mathbf{F}_{\text{lang}}(\mathbf{x}, s)$, where $\mathbf{x}$ is the 3D position and $s$ is the scale.

To supervise this field, LERF uses a precomputed multi-scale feature pyramid of CLIP embeddings as ground truth. The feature pyramid is generated from patches of input multi-view images at different scales. Utilizing volumetric rendering \cite{volumetric_rendering}, the language field can be used to render CLIP embeddings in 2D along each ray $\vec{\mathbf{r}}(t) = \mathbf{o} + t\mathbf{d}$:
\begin{equation}\label{eq1}
\phi_{\text{lang}}(\mathbf{r}) = \int T(t) \sigma(t) \mathbf{F}_{\text{lang}}(\mathbf{r}(t), s(t)) \, dt,
\end{equation}where $T(t)$ represents accumulated transmittance, $\sigma(t)$ is the volume density, and $s(t)$ adjusts according to the distance from the origin, enabling efficient, scale-aware 3D relevance scoring. The rendered CLIP embedding is then normalized to the unit sphere similar as in \cite{clip}. 

The main objective during training is to align the rendered CLIP embeddings with the ground truth CLIP embeddings by minimizing the following loss:
\begin{equation}\label{eq2}
    \mathcal{L}_{\text{lang}} = -\lambda_{\text{lang}} \sum_i \phi_{\text{lang}} \cdot \phi_{\text{gt}}
\end{equation}where $\phi_{\text{lang}}$ denotes the rendered CLIP embedding, $\phi_{\text{gt}}$ is the corresponding target embedding from the precomputed feature pyramid, and $\lambda_{\text{lang}}$ is the loss weight. This loss encourages CLIP embeddings in the language field to align with its language-driven semantic features, thereby allowing open-vocabulary queries within the 3D scene.

\subsection{Geometric primitive}

A core component of SPNeRF is the introduction of geometric primitives. Following recent works \cite{yin2024sai3dsegmentinstance3d,yang2023sam3dsegment3dscenes}, we employ a normal-based graph cut algorithm \cite{Felzenszwalb2004EfficientGI} to over-segment the point cloud \( P \in \mathbb{R}^{N \times 3} \) into a collection of superpoints \( \{ Q_i \}_{i=1}^{N_Q} \), this results in higher-level groupings that better capture the geometric structure of the scene. By aggregating CLIP features at the superpoint level rather than for individual points, we produce more coherent representations, addressing the ambiguities often encountered with point-wise embeddings.

To ensure consistency in the aggregated CLIP features and to align the NeRF representation with the input point cloud, we introduce two complementary loss functions: a \textit{consistency loss} and a \textit{density loss}.

\paragraph{Consistency Loss.} To promote consistency across batches of points within superpoints, we employ the consistency loss on sampled pairs of point embeddings following the Huber loss, then average the results across multiple scales. This loss makes the CLIP embeddings more resilient to outliers, allowing the embeddings to align closely with the majority of points in each batch. Given two embeddings, \( \mathbf{f}_i \) and \( \mathbf{f}_j \), from a batch of sampled points within a superpoint, the consistency loss for each pair is defined as:

\begin{equation}\label{eq3}
\mathcal{L}_{\text{c}}(\mathbf{f}_i, \mathbf{f}_j) = 
\begin{cases} 
\frac{1}{2} \|\mathbf{f}_i - \mathbf{f}_j\|^2 & \text{if } \|\mathbf{f}_i - \mathbf{f}_j\| \leq \delta \\
\delta \|\mathbf{f}_i - \mathbf{f}_j\| - \frac{1}{2} \delta^2 & \text{if } \|\mathbf{f}_i - \mathbf{f}_j\| > \delta
\end{cases}
\end{equation}where \( \delta \) is a threshold parameter that determines the transition between the quadratic and linear regions of the loss. The overall consistency loss for a batch is then averaged across all scales as follows:

\begin{equation}\label{eq4}
\mathcal{L}_{\text{c\_batch}} = \frac{1}{N} \sum_{(i, j) \in \textit{batch}} \frac{1}{S} \sum_{k=1}^{S}  \mathcal{L}_{\text{c}}(\mathbf{f}_{i_k}, \mathbf{f}_{j_k})
\end{equation} where \( N \) is the number of sampled point pairs, and \( \textit{batch}\) represents the set of sampled pairs within the superpoints, \( S \) is the number of scales. This averaging across scales encourages consistent feature alignment within superpoints.

\paragraph{Density Loss.} To ensure that NeRF accurately captures the geometry of the 3D scene, we use a density loss to guide NeRF's density field based on the point cloud positions. For a given point $p_i$ from the point cloud, we encourage the NeRF density $\sigma(p_i)$ at that location to be close to 1, indicating high occupancy:

\begin{equation}\label{eq5}
\mathcal{L}_{\text{density}} = \frac{1}{N} \sum_{i=1}^{N} \left( 1 - \sigma(p_i) \right)^2
\end{equation} this loss ensures that NeRF correctly represents the occupied regions of the 3D space, aligning the density field with the underlying point cloud geometry.

\paragraph{Progressive Training.} To ensure effective optimization of SPNeRF, we employ a progressive training strategy. Initially, we apply the NeRF color rendering loss \cite{mildenhall2020nerf} during training to allow the geometry to converge and establish an spatial structure. Then, we introduce the CLIP language embedding loss $\mathcal{L}_{\text{lang}}$, enabling the language field to learn meaningful language features for positions in the 3D field. Finally, we incorporate the consistency loss $\mathcal{L}_{\text{c\_batch}}$ and density loss $\mathcal{L}_{\text{density}}$ to enhance the consistency and robustness of the CLIP embeddings within superpoints. This staged training process ensures a balanced and efficient optimization of both the geometric and semantic components of SPNeRF.

\subsection{Merging block}

Instead of relying on per-point clustering, SPNeRF assigns segmentation labels based on the relevancy score between superpoint-level CLIP embeddings and the target class label embeddings. 

\paragraph{Relevancy Score.} After training, we begin by using farthest point sampling to select \( N_p \) representative points within each superpoint. Given a sampled point \( p_i \) in the superpoint \( sp_n \), \( i \in \{1, 2, \dots, N_p\} \), with \( N_p \) being the number of sampled points, \( n \in \{1, 2, \dots, N\} \) and $N$ the number of superpoints. We retrieve the CLIP embedding \( \mathbf{f}_{p_i} \) of point $p_i$ by querying the SPNeRF CLIP feature field at \( p_i \)'s position. These embeddings collectively represent the superpoint \( sp_n \)'s CLIP feature set.

Next, the target class label is encoded using the CLIP text encoder to produce a positive CLIP embedding \( \mathbf{f}_{\text{pos}} \). As proposed by LERF \cite{lerf2023}, we also define a set of negative CLIP embeddings \(\{ \mathbf{f}_{\text{neg}_k}\}\), \(k \in \{1, 2, \dots, K\} \).  \( \mathbf{f}_{\text{neg}_j} \) represent the encoded features of canonical text like \textit{"object"} and \textit{"things"}. For each representative embedding \( \mathbf{f}_{p_i} \) within the superpoint, we compute the cosine similarity with both the positive class embedding and each negative embedding. For each point \( p_i \), its relevancy score $R_{p_i}$ is the minimum score across all negative comparisons after softmax normalization:
\begin{equation}\label{eq6}
    F(\mathbf{f}_1, \mathbf{f}_2) = \exp(\text{sim}(\mathbf{f}_1, \mathbf{f}_2))
\end{equation}
\begin{equation}\label{eq7}
R_{p_i} = \min_K \left( \textstyle \frac{F(\mathbf{f}_{p_i}, \mathbf{f}_{\text{pos}})}{F(\mathbf{f}_{p_i}, \mathbf{f}_{\text{pos}}) + F(\mathbf{f}_{p_i}, \mathbf{f}_{\text{neg}_k})} \right)
\end{equation} where exp is the exponential function, $\text{sim}$ is the cosine similarity, and $K$ is the number of negative embeddings.

The relevancy score $R_{sp_n}$ for a given superpoint \( sp_n \) is the median relevancy scores of all sampled points, which is robust against outliers. The median point's CLIP embedding is also used to represent the superpoint's CLIP embedding \( \mathbf{f}_{sp_n} \) .

\begin{figure*}[t]
  \centering
  \includegraphics[width=\textwidth]{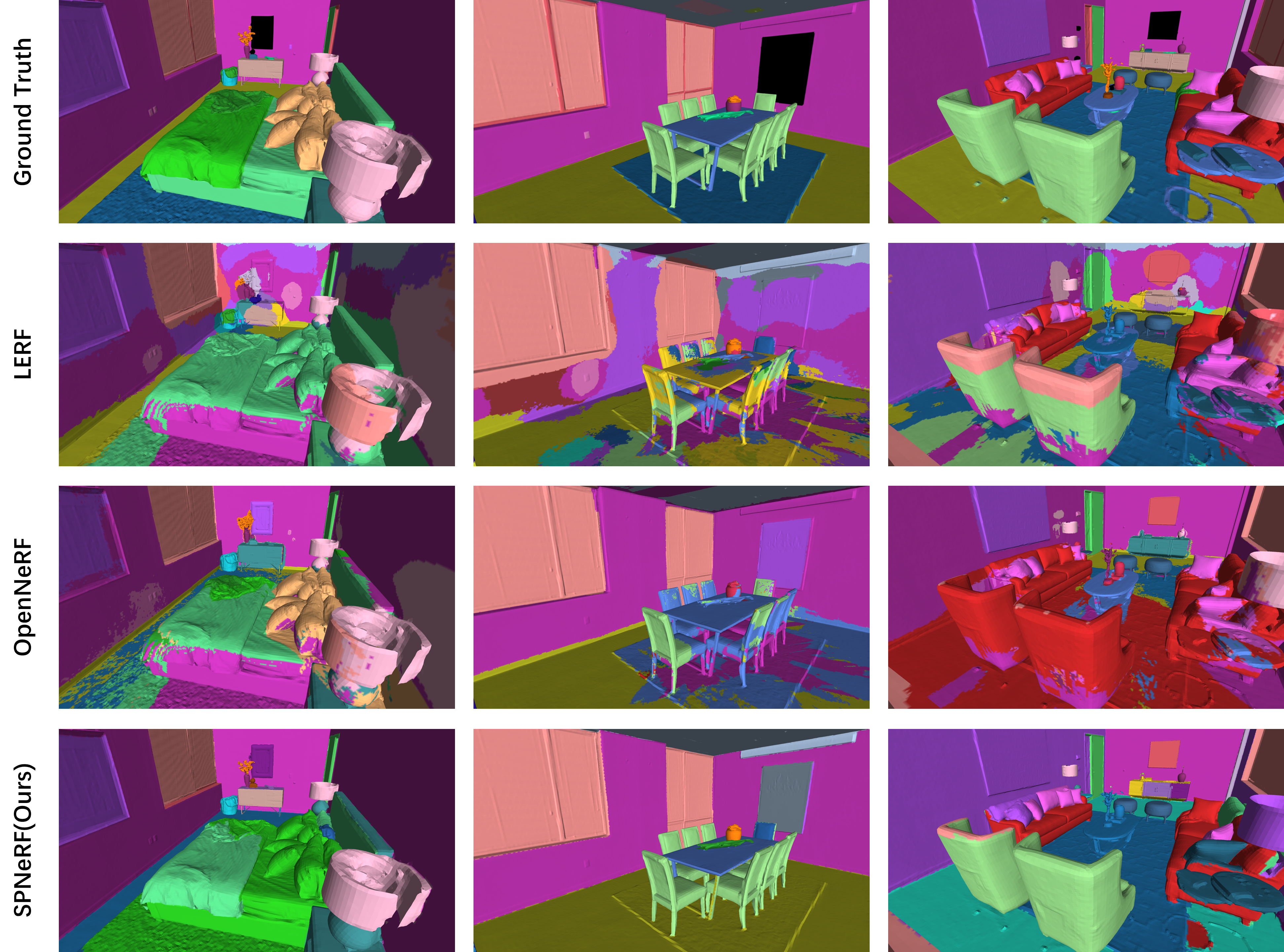}
  \caption{\textbf{3D segmentation results in comparison to other methods.} Qualitative comparison of 3D semantic segmentation results on the Replica dataset. Rows display results from (top to bottom) ground truth, LERF, OpenNeRF and SPNeRF, across 3 indoor scenes. SPNeRF demonstrates improved boundary coherence and segmentation accuracy in general.}
  \label{fig:results_overview}
 \end{figure*}
 
\paragraph{Affinity Score.} To further enhance the effect of relevancy score, we introduce an affinity score. Same as calculating the relevancy score, we need positive and negative embeddings for comparison to define the score. For the given class, we choose \(N_{a}\) superpoints which have the highest relevancy score as positive superpoints $sp_{pos_j}$, and use their CLIP embeddings as positive embeddings\(\{ \mathbf{f}_{\text{pos}_j}\}\), \(j \in \{1, 2, \dots, N_{a}\}\). We choose another \(N_{a}\) superpoints which have the lowest relevancy score as negative superpoints $sp_{neg_k}$, and use their CLIP embedding as negative embeddings \(\{ \mathbf{f}_{\text{neg}_k}\}\), \(k \in \{1, 2, \dots, N_{a}\}\). In order to calculate the affinity score $A_{sp_n\_sp_{pos_j}}$ between a superpoint \( sp_n \) and a positive superpoint $sp_{pos_j}$, we compare \( \mathbf{f}_{sp_n} \) with each positive embedding $\mathbf{f}_{\text{pos}_j}$ and the set of $N_{a}$ negative embeddings \(\{ \mathbf{f}_{\text{neg}_j}\}\), and select the minimum score across all negative comparisons:

\begin{equation}\label{eq8}
A_{sp_n\_sp_{pos_j}} = \min_{N_a} \left( \textstyle \frac{F(\mathbf{f}_{sp_n}, \mathbf{f}_{\text{pos}_j})}{F(\mathbf{f}_{sp_n}, \mathbf{f}_{\text{pos}_j}) + F(\mathbf{f}_{sp_n}, \mathbf{f}_{\text{neg}_k})} \right)
\end{equation}

Then, we use the relevancy score $R_{pos_j}$ of each positive superpoint $sp_{pos_j}$ as weight to average all $N_a$ affinity scores, and acquire the affinity score $A_{sp_n}$ for the superpoint $sp_n$:
\begin{equation}\label{eq9}
A_{sp_n} = \frac{\sum_{j=1}^{N_a} R_{pos_j} \cdot A_{sp_n\_sp_{pos_j}}}{N_a}
\end{equation}

The relevance score $R_{sp_n}$ for superpoint $sp_n$ is then scaled with affinity. The scaled relevancy score $R_{sp_n}^*$ can be calculated as:
\begin{equation}\label{eq10}
R_{sp_n}^* = R_{sp_n} \cdot w \cdot (1 + (A_{sp_n} -  \min_{N} \left(A_{sp_n}\right))
\end{equation}where $w$ is the affinity weight, and N is the number of superpoints. 

For each superpoint $sp_n$,  the class with highest scaled relevancy score is assigned during semantic segmentation.

\section{\uppercase{Experiments}}
In this section, we present our experimental evaluation assessing both quantitative and qualitative performance. We compare its performance in zero-shot 3D segmentation with respect to the baseline methods LERF and OpenNeRF. In addition, we conduct an ablation study to analyze the contribution of each of SPNeRF’s components, including the consistency loss and affinity alignment.

\begin{figure*}[t]
  \centering
  \includegraphics[width=\textwidth]{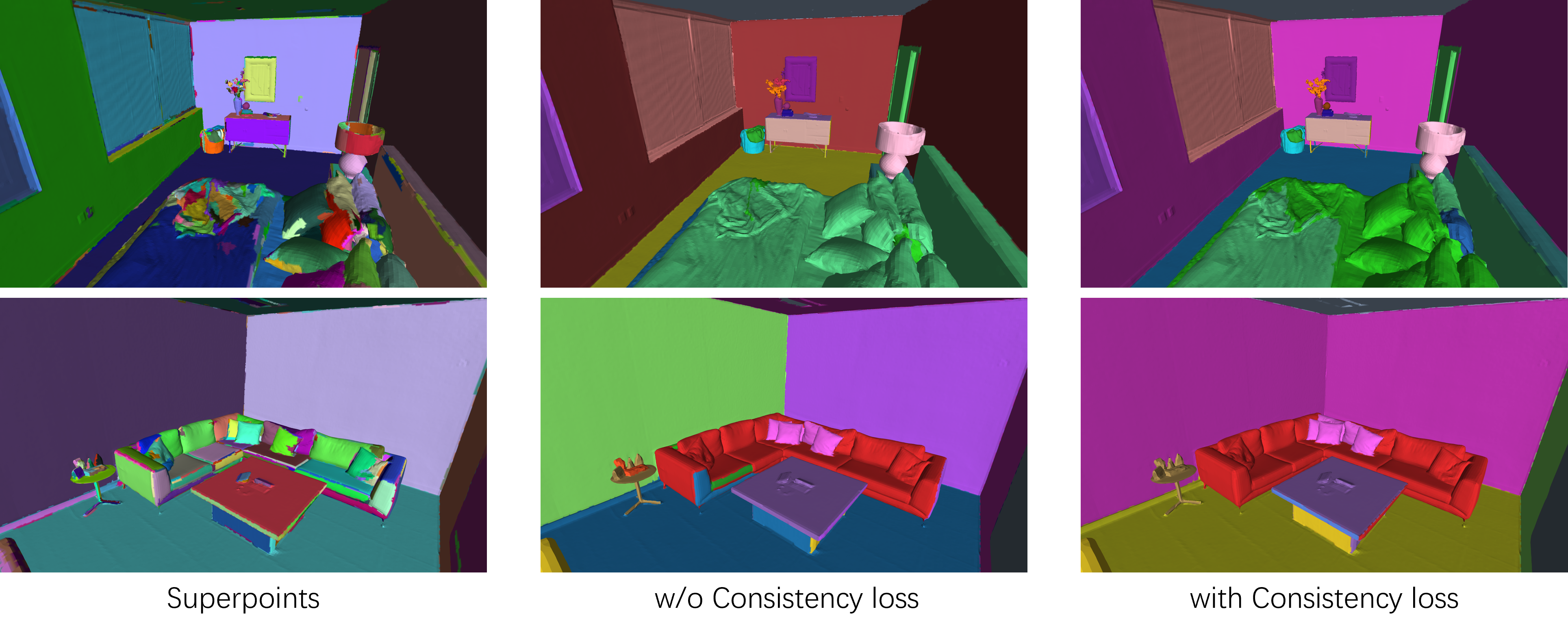}
  \caption{\textbf{Ablation comparison of consistency loss.} Even contrained by the fragmented superpoints, the results w/o loss tend to be consistent due to the image embedding characteristic of CLIP. The consistency loss helps the model to get more precise semantic info, especially for large superpoints like wall surfaces.}
  \label{fig:ablation1}
 \end{figure*}

\subsection{Experiment Setup}

We evaluated SPNeRF on the Replica dataset, a standard benchmark for 3D scene understanding. Replica dataset comprises photorealistic indoor scenes with high-quality RGB images and 3D point cloud data, annotated with per-point semantic labels for a variety of object categories. This dataset serves as a robust benchmark for evaluating segmentation in complex, densely populated indoor environments. For each scene in Replica, 200 posed images are used for all experiments.

\begin{figure*}[t]
  \centering
  \includegraphics[width=\textwidth]{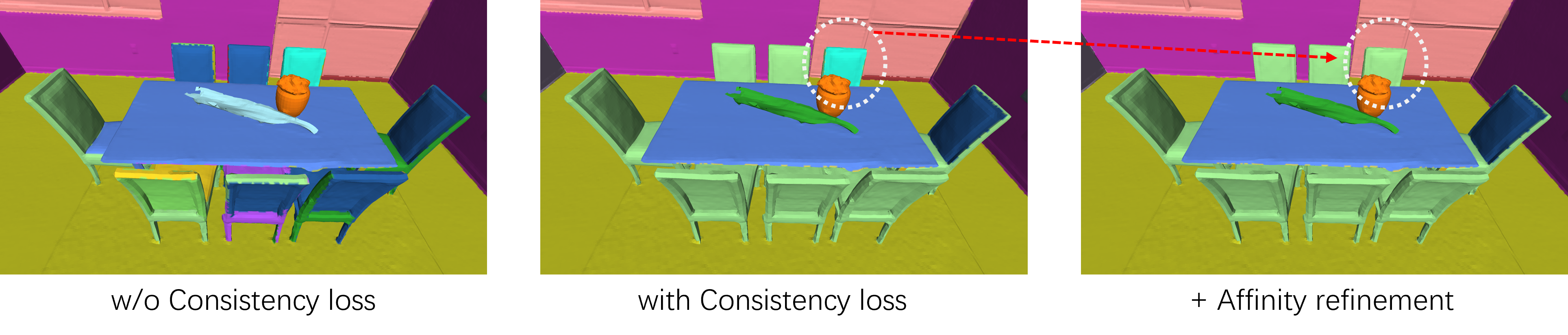}
  \caption{\textbf{Ablation comparison.} The figure illustrates the improvement by consistency loss and affinity score.}
  \label{fig:ablation2}
 \end{figure*}

For image language features extraction, we utilized OpenCLIP ViT-B/16 model. We trained our SPNeRF with posed RGB images and 3D geometry, applied zero-shot semantic segmentation without additional fine-tuned or pre-trained 2D segmentation models. For evaluation, we followed the approach of \cite{peng2023openscene3dsceneunderstanding}. The accuracy of the predicted semantic labels is evaluated using mean intersection over union (mIoU) and mean accuracy (mAcc).

\subsection{Method Comparison}
\paragraph{Quantitative evaluation.} We compare with OpenNeRF \cite{engelmann2024opennerf}, OpenScene \cite{peng2023openscene3dsceneunderstanding} and LERF \cite{lerf2023}. To evaluate LERF, we generated segmentation masks by rendering relevancy maps, projecting them onto Replica point clouds, and assigning each point of the class with the highest score, same as \cite{engelmann2024opennerf} proposed. OpenNeRF is evaluated with their provided code. In order to provide a comparison of the models' own effectiveness on segmentation, we did not use NeRF-synthesized novel views to fine-tune the models during comparison. In contrast, SPNeRF and LERF use RGB images as input, while OpenNeRF takes RGB images and corresponding depth maps as input. Results of OpenScene are taken from \cite{engelmann2024opennerf}.
\vspace{0.5em}
\begin{table}[ht]
\caption{Quantitative results on Replica dataset for 3D semantic segmentation.}
\centering
\begin{tabular}{lcc}
\hline
Method & mIoU & mAcc \\
\hline
LERF & 10.5 & 25.8 \\
OpenScene  & 15.9 & 24.6 \\
OpenNeRF & 19.73 & 32.61 \\
SPNeRF (Ours) & 17.25 & 31.07 \\
\hline
\end{tabular}
\label{tab:replica_comparison}
\end{table}

Table \ref{tab:replica_comparison} summarizes the 3D semantic segmentation performance of SPNeRF relative to baseline methods on the Replica dataset. SPNeRF achieves competitive scores with a mIoU of 17.25\% and mAcc of 31.07\%. While LERF and SPNeRF both use original CLIP to extract semantic information,  SPNeRF improves significantly over the baseline LERF. Although OpenNeRF attains the highest overall performance with the support of a fine-tuned 2D model for segmentation, SPNeRF's results emphasize its effective integration of superpoint-based feature aggregation and language-driven embeddings without additional 2D segmentation knowledge.

The experimental results demonstrate SPNeRF's enhanced capability in maintaining feature consistency within superpoints, especially when evaluated against the LERF baseline. The 6.75\% improvement in mIoU over LERF without structural network changes illustrates the impact of our approach in aligning semantic language features spatially across 3D fields. Without any 2D segmentation knowledge, SPNeRF's results align closely quantitatively with OpenNeRF which is trained with a 2D segmentation model, indicating CLIP's potential for fine-grained segmentation.

\paragraph{Qualitative evaluation.} Figure \ref{fig:results_overview} illustrates a qualitative comparison of segmentation results between SPNeRF, OpenNeRF, and LERF across various indoor scenes in the Replica dataset. While the other methods' segmentation tend to splash near boundaries, SPNeRF demonstrates great boundary coherence and spatial consistency, particularly in scenes with complex object arrangements. OpenNeRF, while generally robust in correct class estimation, exhibits minor loss of detail in cluttered environments. SPNeRF’s superpoint-based segmentation mitigates these issues by aggregating features within geometric boundaries, resulting in coherent representations, especially when comparing the wall areas with LERF, SPNeRF learns to concentrate on the correct semantic even using same network structure.

Overall, the qualitative results highlight the ability of SPNeRF to deliver competitive 3D segmentation in complex scenes, complementing its quantitative gains in mIoU and mAcc. The combination of CLIP embeddings and superpoint-based relevancy scoring enables SPNeRF to differentiate structures and maintain semantic consistency across object boundaries, reducing noise and improving clarity in less visible areas.

\subsection{Ablation Study}
To analyze the impact of SPNeRF’s individual components, we perform an ablation study by systematically removing the primitive consistency loss and affinity-based refinement. Table~\ref{tab:ablation} presents the quantitative results. Removing the primitive consistency loss results in a notable decrease in mIoU (from 17.25 to 15.31) and mAcc (from 31.07 to 26.82), highlighting its importance in preserving coherent embeddings within superpoints. As also shown in Figure \ref{fig:ablation1}, consistency loss largely improved the precision of classification, especially for large superpoints like walls, which are more likely to contain different semantic embeddings, consistency loss helps the CLIP feature field to learn the most important and distributed semantics of superpoints. Similarly, excluding the affinity-based refinement slightly reduced the performance numerically. As shown in Figure \ref{fig:ablation2}, affinity refinement can improve the segmentation quality by capturing semantic relationships between superpoints, for example chair surfaces, and maintain the possibility to over cover adjacent parts.

\begin{table}[ht]
\caption{Ablation study results on the Replica dataset. Both the primitive consistency loss and affinity refinement contribute significantly to SPNeRF’s overall segmentation quality.}
\centering
\begin{tabular}{lcc}
\hline
Model Variant & mIoU & mAcc \\
\hline
Full SPNeRF & 17.25 & 31.07 \\
w/o Affinity Refinement  & 17.13 & 30.77 \\
w/o Consistency Loss& 15.31 & 26.82 \\
w/o both & 13.78 & 24.59\\
\hline
\end{tabular}
\label{tab:ablation}
\end{table}

\section{\uppercase{Conclusion}}
\label{sec:conclusion}

We introduced SPNeRF, a zero-shot 3D segmentation approach that enhances Neural Radiance Fields (NeRF) through the integration of geometric primitives and visual-language features. Without training on any ground truth labels, our model can semantically segment unseen complex 3D scenes. By embedding superpoint-based geometric structures and applying a primitive consistency loss, SPNeRF overcomes the limitations of CLIP’s image-based embeddings, achieving higher spatial consistency and segmentation quality in 3D environments, while mitigating ambiguities in point-wise embeddings. SPNeRF outperforms LERF and performs competitively with OpenNeRF, while SPNeRF avoids additional 2D segmentation models required by OpenNeRF. While SPNeRF has demonstrated competitive performance, it also inherits limitations from CLIP’s 2D image-based embeddings, leading to occasional ambiguities in details. Future work could explore more efficient alternatives to NeRF, such as Gaussian splatting \cite{kerbl3Dgaussians} or efficiently incorporating 2D foundation models like the Segment Anything Model (SAM) \cite{sam} to enable instance-level segmentation.

\section*{\uppercase{Acknowledgements}}
This work has partly been funded by the German Federal Ministry for
Digital and Transport (project EConoM under grant
number 19OI22009C).

\bibliographystyle{apalike}
{\small
\bibliography{main}}

\end{document}